\documentclass[conference]{IEEEtran}
\ifCLASSINFOpdf
\else
\fi
\usepackage{subfigure}
\usepackage{url}
\usepackage{acronym}
\usepackage{subfig}
\usepackage{booktabs}
\usepackage{siunitx}
\usepackage{graphicx}
\usepackage{tikz}
\usetikzlibrary{calc}
\usetikzlibrary{shadows}
\usetikzlibrary{shapes}
\usepackage{amsmath}
\usepackage{multirow}
\usepackage[rotateright]{rotating}
\usepackage{acronym}
\usepackage{subfigure}
\usepackage{epstopdf}
\newcommand{\bi}{\begin{itemize}}\newcommand{\ei}{\end{itemize}}
\newcommand{\be}{\begin{enumerate}}\newcommand{\ee}{\end{enumerate}}
\newcommand{\bs}{\begin{slide}}\newcommand{\es}{\end{slide}}
\newcommand{\bc}{\begin{center}}\newcommand{\ec}{\end{center}}
\newcommand{\beq}{\begin{equation}}\newcommand{\eeq}{\end{equation}}
\newcommand{\beqn}{\begin{eqnarray}}\newcommand{\eeqn}{\end{eqnarray}}
\newcommand{\btab}{\begin{tabular}}\newcommand{\etab}{\end{tabular}}

\acrodef{mAP} {Mean Average Precision}
\acrodef{IoU} {Intersection Over Union}
\acrodef{BoW} {Bag of Words}
\acrodef{FRCN}{Fast Region-based Convolutional Network }
\begin{document}
%
\title{
\vspace*{-0.75cm}{\Large Accepted as a Conference Paper for IJCNN 2016}\\
Automatic Graphic Logo Detection via Fast Region-based Convolutional Networks}

\author{\IEEEauthorblockN{Gon\c{c}alo Oliveira\IEEEauthorrefmark{1},
Xavier Fraz\~{a}o\IEEEauthorrefmark{2},
Andr\'{e} Pimentel\IEEEauthorrefmark{2}, 
Bernardete Ribeiro\IEEEauthorrefmark{3}}
\IEEEauthorblockA{\IEEEauthorrefmark{1}Center for Informatics and Systems of the University of Coimbra\\
Polo II, 3030-290 Coimbra, Portugal,
\\ Email:goncalopalaio@gmail.com}
\IEEEauthorblockA{\IEEEauthorrefmark{2}EyeSee Solutions\\
Av 5 Outubro 293, 1600-035 Lisbon, Portugal\\
Email: \{xavier.frazao,andre.pimentel\}@eyesee.pt}
\IEEEauthorblockA{\IEEEauthorrefmark{3}CISUC - Department of Informatics Engineering\\
University of Coimbra, Portugal\\
Email: bribeiro@dei.uc.pt}}
\maketitle

\begin{abstract}
Brand recognition is a very challenging topic with many useful applications in localization recognition, advertisement and marketing.
In this paper we present an automatic graphic logo detection system that robustly handles unconstrained imaging conditions. Our approach is based on Fast Region-based Convolutional Networks (FRCN) proposed by Ross Girshick, which have shown state-of-the-art performance in several generic object recognition tasks (PASCAL Visual Object Classes challenges). In particular, we use two CNN models pre-trained with the ILSVRC ImageNet dataset and we look at the selective search of windows `proposals' in the pre-processing stage and data augmentation to enhance the logo recognition rate. The novelty lies in the use of transfer learning to leverage powerful Convolutional Neural Network models trained with large-scale datasets and repurpose them in the context of graphic logo detection. Another benefit of this framework is that it allows for multiple detections of graphic logos using regions that are likely to have an object. Experimental results with the FlickrLogos-32 dataset show not only the promising performance of our developed models with respect to noise and other transformations a graphic logo can be subject to, but also its superiority over state-of-the-art systems with hand-crafted models and features.
\end{abstract}
%
\IEEEpeerreviewmaketitle
\section{Introduction}
\label{sec:intro}
Advertisement is the primary way to get revenue or gathering interest for a product or service. Most of the current solutions use implicit placement of ads without regard of context or content, which has been shown to be an effective way of raising brand awareness. In order to well target the user, there is a need to analyze the content and relevant advertisements that exist nowadays.
In order to captivate their customers and make better decisions, companies have the need to analyze the presence of their brand in images and other types of content.
Brand logos help to assess the identity of something or someone. Most solutions use graphic logos as the main target for brand detection since they often present distinct shapes and appear in high contrast regions. It is a challenge since they are often subject to multiple angles and sizes, varying lighting conditions and noise. 
Therefore, although they give perceptual distinctiveness, it is a challenge due to all the conditions they can appear in. 
Most previous works in this context have been based considerably on SIFT \cite{Lowe2004}. This method provides representations and transformations to image gradients that are invariant to affine transformations and robust when facing lighting conditions and clutter. They can also detect stable salient points in the image across multiple scales, usually called key-points.
These previous works build models upon these representations to better capture specific patterns present in graphic logos. For instance, in~\cite{RombergICMR2011}, a shape representation built with found key-points and their respective SIFT representation is proposed for scalable logo recognition in images. Similarly,  in~\cite{romberg2013bundle}  bundles of SIFT features are built from local regions around each key-point to index specific graphic logo patterns.  Although there are many successful works covering a broad range of methods that have been applied in the past, recently, we assisted to the blossom of Convolution Neural Networks (CNNs) in the area of computer vision. 
CNNs have been producing an impressive impact on image recognition. CNNs are composed by many layers which resemble the simple and complex cells in the primary visual cortex~\cite{Wiesel59}. Their structure is hierarchical and are specifically designed to recognize visual patterns directly from image pixels. The convolutional layers alternate with subsampling layers which may vary in number and size. The works in ~\cite{Fukushima80}~\cite{Fukushima2003},~\cite{Schmidhuber96} and~\cite{lecun1998gradient} have been pioneers  for the current CNNs that are researched today. Recently CNNs have been in the center of object recognition research. The rekindled interest in CNNs is largely attributed to \cite{krizhevsky2012imagenet} CNN model, that showed significantly higher image classification accuracy on the 2012 ImageNet Large Scale Visual Recognition Challenge (ILSVRC)~\cite{ILSVRC15}. Their success resulted from a model inspired by LeCun previous work and a few twists that enabled training with 1.2 million labeled images (\textit{e.g.} GPU programming, max($x$,0) rectifying non-linearities and \textit{dropout} regularization).
Our work is focused on providing a way towards graphic logo detection by utilizing general region proposal algorithms and state-of-the-art object recognition systems. However, due to the lack of a large scale dataset with such graphic logos, training a modern system with CNNs from the scratch is mostly unfeasible, therefore, we use transfer learning and data augmentation to ameliorate this problem.

The remaining of this paper is organized as follows. Section~\ref{sec:approach} emphasizes the contributions for automatic logo recognition with deep convolutional neural networks.  Section \ref{sec:transfer} focuses on the transfer learning methodology which apart from fine tuning of parameters enables less costly architectures. 
Section~\ref{sec:Fast} describes the architecture for object detection dubbed as Fast Region-based Convolutional Networks (FRCN). Section~\ref{sec:experimental} presents the research design, the metrics and the discusses the performance results.
Finally, in Section~\ref{sec:conclusion} we address the main contributions of this paper and propose improvements that can be further explored. 
\section{Our approach and contributions}
\label{sec:approach}
We extend and build upon general concepts and models of object recognition in images. Thus, we propose a solution that takes advantage of specific characteristics of graphic logos based on current cutting-edge research.
We focus on company graphic logos since they give a distinctive way of assessing a particular advertising campaign or brand.
The system we propose uses transfer learning to leverage image representations learned with CNNs on large-scale annotated datasets. The transferred representation leads to significantly improved results for brand detection. This is  very important in this study and we empirically show that it performs well and can be as well applied in other contexts and problems.
\section{Transfer learning with CNNs}
\label{sec:transfer}
The convolutional layer of a CNN consists of a set of learnable filters that activate with specific image features.
Earlier layers of these networks learn to detect generic features like edges, and as we move into further layers, the learned features begin to model high level concepts and get more and more specific towards the original dataset. It is then possible to continue training and re-purpose, or transfer the weights so that they adapt to other datasets. Several authors have shown that fine-tuning these networks and transferring features, even from tasks that are not closely similar, can be advantageous when compared to training from the scratch \cite{chatfield2014return}. The process takes advantage of the more generic learned features, given that both datasets `live' in a similar domain, in this case, object detection in images. This method enables training a large network with a small dataset without overfitting. 
\section{Fast Region-based Convolutional Networks (FRCN)}
\label{sec:Fast}
Graphic logos are not usually the main focus of an image, so they are often present in small sizes and partially occluded. Performing classification using the full image would introduce high amounts of background noise. Therefore, performing a more exhaustive search is required in this context. To prevent such an exhaustive search recent methods  employ region `proposals' to reduce the initial search space and perform detection using powerful CNN models~\cite{frcn_girshick2015fast}.
\begin{figure}[htb]
\begin{center}
\includegraphics[width=0.485\textwidth]{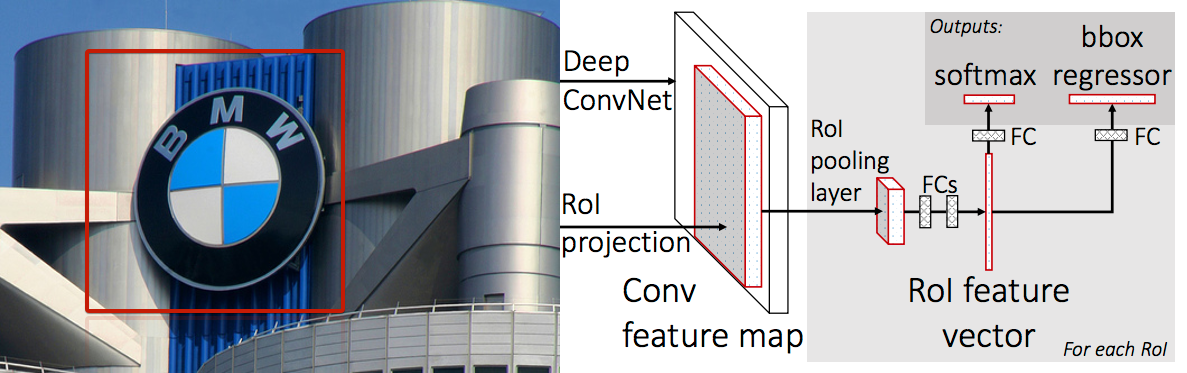}
\end{center}
   \caption{FRCN model. Adapted from \cite{frcn_girshick2015fast}.}
\label{fig:frcn}
\end{figure}
Ross Girshick successfully proposed in~\cite{frcn_girshick2015fast} a single-stage object detection model by using a region proposal algorithm and CNN features. This method performs object detection by classifying multiple regions in the image. This method has shown state-of-the-art performance in PASCAL VOC 2012~\cite{everingham2010pascal}. Its high level representation is shown in Figure \ref{fig:frcn} slightly adapted from~\cite{frcn_girshick2015fast}.
Using category-independent region `proposals', the model performs classification in a reduced set of regions that are likely to have present an object; this helps to reduce the potential false-positives and the number of regions to analyze. In particular, it uses the approach in~\cite{uijlings2013selective} based on a selective search for object recognition. This method generates regions using clustering of similar regions and several diversification methods, by capturing possible regions where an object might be, simply without the use of sliding windows. 
Feature extraction is performed using a CNN for each one of the proposed regions. These regions were then classified using a softmax classifier with a Fully Connected (FC) layer and post-processed using non-maxima suppression. After classification priority is given to regions with a high confidence value (local maxima). Duplicate region are iteratively removed with non-maximum suppression. Moreover, localization of the object is further refined using bounding-box regressors. 
\section{Experimental Results}
\label{sec:experimental}
In order to assess our approach and the validity of our statements we conducted a set of experiments on a logo benchmark dataset. We provide further details of the dataset, indicate the performance measures, describe the augmentation pre-processing procedure, present  and discuss the results.
\subsection{Experimental Setup}
We evaluate the FRCN model for brand detection with the FlickrLogos-32 dataset~\cite{dataset_website} which contains 32 graphic logo class images. The training set contains 320 images, the validation set 960 and the test set 960 images, each of which showing at least one single logo. Additionally, the test set contains 3000 images that do not contain any logo from the 32 classes.  A detailed characterization of the dataset is given in Table~\ref{table:flickr32_dist}. 
\begin{table}[htb]
\caption{Flickr32 Logo Dataset\cite{RombergICMR2011} partitions/subsets. Slightly adapted from \cite{dataset_website}.}
\centering
\begin{tabular}{p{4em}|p{13em}|l|l}
Partition & Description & Images/class & \#Images \\
\hline
Training set & Hand-picked images & 10 per class & 320 \\
\hline
Validation set & Images showing at least a single logo under various views & 30 per class & 960  \\
& Non-logo images & 3000 & 3000 \\
\hline
Test set & Images showing at least a single logo under various views & 30 per class & 960\\
& Non-logo images & 3000 & 3000 \\\hline
\end{tabular}
\label{table:flickr32_dist}
\end{table}
The dataset is annotated with image binary masks. Testing is performed in images that do not only contain graphic logos, but also some amount of background noise.
\begin{figure}[htb]
\includegraphics[width=1.0\linewidth]{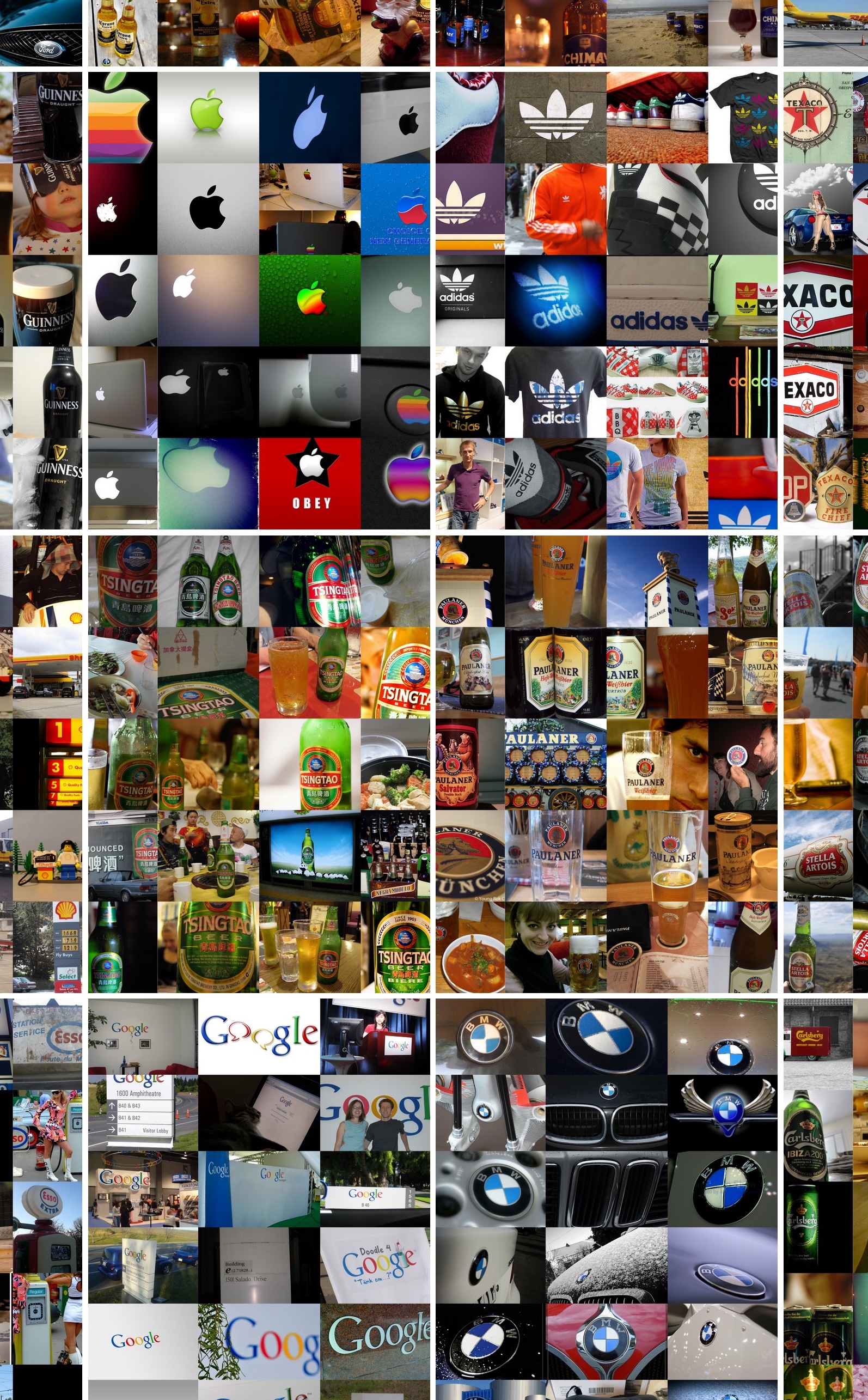}\\
\caption{Graphic Logos, FlickrLogos-32 dataset \protect\cite{RombergICMR2011}}
\label{fig:dataset}
\end{figure}
\subsection{Model Evaluation}

\paragraph{Intersection Over Union}

As mentioned in Section~\ref{sec:Fast}, the method from\cite{uijlings2013selective} is used to generate region `proposals' for each image during training. The \acf{IoU} metric is used to evaluate which regions should be considered as a positive sample given the region bounding box overlap with the ground-truth object.

This metric is often used to judge correct object detections in images. Detections usually are considered correct when the area of overlap $a_0$ between the predicted bounding box and ground-truth exceeds 50\% \cite{everingham2010pascal} using Equation \ref{eq:iou}.

\begin{equation}
	a_0 = \frac{area \left( B_p \cap B_{gt} \right)}{ area \left( B_p \cup B_{gt} \right)}
\label{eq:iou}
\end{equation}

\paragraph{Non-maximum Suppression}

Since we are classifying multiple region `proposals' in each testing image, many of the detected regions will be heavily overlapped, not only with the ground-truth object, but also with each other. Non-maximum Suppression is a technique that has been used in the edge thinning in the Canny edge detection algorithm \cite{rcnn_GirshickDDM13}. In this context it is used to remove redundant regions that have a low score, but also merging multiple regions that indicate the same object that are given an high score by the classifier.

\paragraph{Mean Average Precision}
While precision and recall are single values based on list of unranked predictions, \acf{mAP} is often used in information retrieval. In image context, we look at a specific detection of a logo in an image. Since the retrieved list is ranked, it is desirable to consider the order that the returned images are presented and give more relevance to correct items at the top.
To obtain the value of \ac{mAP}, P (precision) and R (recall) are then calculated at every position in the ranked list of images for each query. Average Precision is defined as:
\begin{equation}
	Average\,\, Precision = \sum_{k=1}^{N} P(k) \cdot \Delta R(k) \nonumber
	\label{eq:avep}
	\end{equation}
where $k$ is the rank in the list of retrieved documents, $N$ the total number of retrieved documents for this query, $P(k)$ the precision value at the $k$ point of the list and $\Delta R(k)$ the change in recall value since the last point in the list.
Mean Average Precision (\ac{mAP}) is the mean value of average precision values obtained across all queries or classes.
The Average Precision also summarizes the shape of the precision/recall curve built from the method's ranked output. Most of the recent computer vision works use this metric for classification and detection.
\subsection{Pre-Processing}
We performed data augmentation by horizontally flipping the training images. Since the training set only contains $10$ images per class we also used the original validation set for training. In addition, as a separate experiment we randomly distort each image with a shear operation ([-5,5] degrees) and apply a slight shift in color values (3 percent) on each image.
The method from \cite{uijlings2013selective} is used to generate region `proposals' for each image in the dataset. Selective Search in~\cite{uijlings2013selective} first applies image segmentation and then performs bottom-up hierarchical grouping of the neighboring segments to generate regions. This step produces a varying number or regions depending on the image characteristics and the diversification methods when grouping segments. Defining different sizes for the initial segments is a way to diversify the produced regions. It is possible to generate regions from multiple initial sizes of segments.
We generate regions using Selective Search in two distinct modes, which we call Fast and Quality mode. These two modes differ on the number of initial segment sizes used for grouping. Quality mode uses four different segment sizes while Fast mode uses only two sizes during the initial grouping. For that reason, quality mode generates a significantly larger and more diversified set of regions than Fast mode.
As mentioned above, using this method will help to reduce background noise and guide the model toward regions where an object is probable to appear in. They are used not only during the testing phase but also during training, where these regions are sampled to fine tune the model.
\subsection{Tools}
We used the Caffe deep learning framework \cite{jia2014caffe,caffe} implementation of the FRCN model~\cite{edison} and present the main results yielded by using two pre-trained CNNs. The first is the Caffe implementation of the CNN described by \cite{krizhevsky2012imagenet} called Caffenet, the second, is a larger model, which has the same depth but wider convolutional layers, VGG{\_}CNN{\_}M{\_}1024, described in \cite{chatfield2014return}. Both were pre-trained with the ILSVRC ImageNet dataset.
We refer to \cite{krizhevsky2012imagenet,chatfield2014return} for more network architecture details.
\subsection{ Setup Summary}
 An overall summary of the experimental setup is given below.
\begin{itemize}
	\item FlickrLogos-32 dataset 
	\begin{itemize}
		\item Real-world images showing brand logos
		\item $32$ different brand logos
		\item $70$ images per class/brand logo
		\item $6000$ non-logo images
	\end{itemize}
	\item CNN models pre-trained with the ILSVRC ImageNet dataset (1000 categories and 1.2 million images)
	\begin{itemize}
		\item Caffenet [2] (BD-FRCN-M$_1$)
		\item VGG{\_}CNN{\_}M{\_}1024 [3] (BD-FRCN-M$_2$)
	\end{itemize}
	\item Caffe Deep Learning Framework
\end{itemize}
\subsection{Results and Discussion}
In this section we first present and discuss the results of the proposed model using \ac{mAP} values. We analyse the detection and recognition F1-scores not only in images with graphic logos, but also in images with no logos.
Convolutional Neural Networks are now considered a strong candidate for visual recognition tasks \cite {razavian2014cnn}. The results we present herein and which will be described in this section, seem to corroborate that fact, moreover if we compare them to our previous efforts~\cite{Oliveira2015}.
A major concern from the beginning was the amount of data available. If we join the training set and validation set of the dataset, $40$ images of each class could be obtained for training. 

As previously mentioned, pre-training a deep network with a similar domain dataset is advantageous. In this case, by using transfer learning we avoid having to build a new large scale dataset for our domain to train a deep network from the scratch.

\begin{figure*}[htb]
\centering
\includegraphics[width=\textwidth]{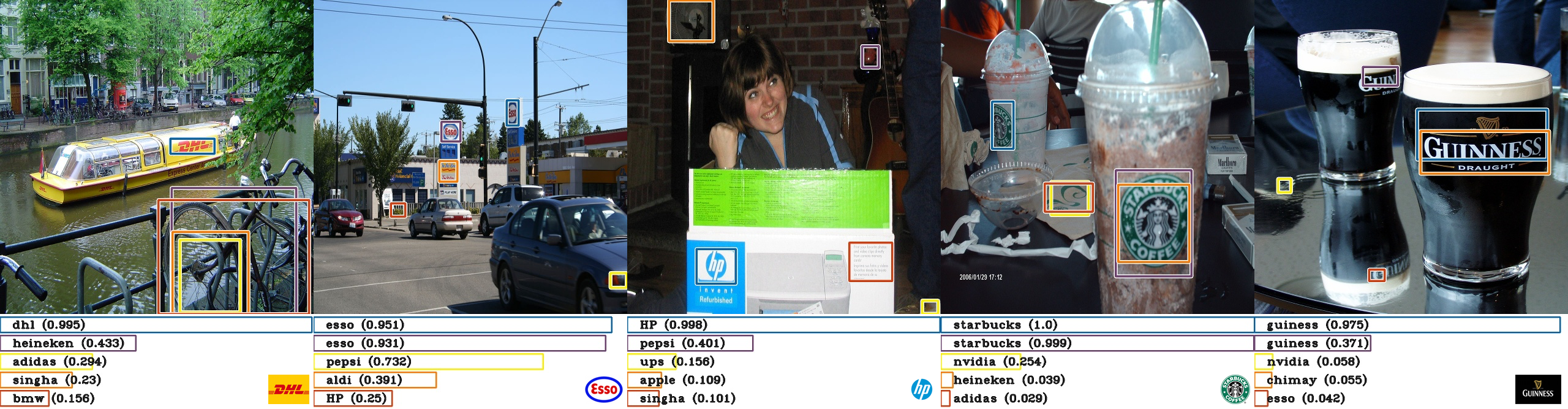}
\caption{Top-5 confidence scores for several example images}
\label{fig:top-5}
\end{figure*}
The flexibility given by the original \ac{FRCN} model is greatly suitable for the brand detection problem. Graphic logos appear in small regions in the image and since the first step is to generate object `proposals', the model gives an intuitive and modern way to solve this issue. Additionally, the model deals directly with background noise by using a special background class and also object `proposals' to reduce the number of regions that are processed in the image.
Moreover, the model is able to detect and localize multiple instances of a graphic logo. As already mentioned, that functionality is helpful in situations where contextual advertisement insertion is needed.
The training process of the FRCN model jointly optimizes a softmax classifier and bounding-box regressors while adapting the CNN weights to our task of graphic logo detection. We start Stochastic Gradient Descent at a small learning rate of $0.001$, which allows fine-tuning of the CNN network to occur without losing the capabilities gained during pre-training. We fine-tune Caffenet and VGG{\_}CNN{\_}M{\_}1024 during $80000$ iterations while saving a snapshot of the models each $10000$ iterations. We use these snapshots to analyze each model performance across the whole fine-tuning process. We alternatively designate the Caffenet model as M$_1$, and the larger model, VGG{\_}CNN{\_}M{\_}1024 as M$_2$. We will abbreviate the name of our final Brand Detection model to BD-FRCN. The specification of the models BD-FRCN-M$_1$ and BD-FRCN-M$_2$ have been above indicated in the Setup Summary.
\begin{table*}
\small\centering
\caption{Comparison of obtained \ac{mAP} values with those of the work in \cite{romberg2013bundle}. Results obtained at 60000 iterations.}
\begin{tabular}{l|l|l|l|l}
Method    & Learning  rate & Selective Search & \ac{mAP} \\  \hline
\ac{BoW} (tf-idf-sqrt, vocabulary-1000)\cite{romberg2013bundle}     &            &                                     & 0.545      \\
Bundle Min Hashing (1p-wgc-ransac) \cite{romberg2013bundle}                               &            &                                & 0.568      \\
BD-FRCN-M$_2$                           & 0.001              & Fast             & 0.7347     \\
BD-FRCN-M$_1$                                  & 0.001              & Fast             & 0.7314     \\
BD-FRCN-M$_2$                        & 0.005              & Fast             & 0.7347     \\
BD-FRCN-M$_1$                               & 0.005              & Fast             & 0.7314     \\
BD-FRCN-M$_2$                        & 0.001              & Quality          & 0.6972     \\
BD-FRCN-M$_1$                                & 0.001              & Quality          & 0.6898     \\
BD-FRCN-M$_2$ - (Shear \& Color)     & 0.001              & Fast             & 0.6941     \\
BD-FRCN-M$_1$ - (Shear \& Color)              & 0.001              & Fast             & 0.6912    
\end{tabular}
\label{table:map_comparison}
\end{table*}

\begin{figure}[htb]
\begin{center}
\includegraphics[width=0.375\textwidth]{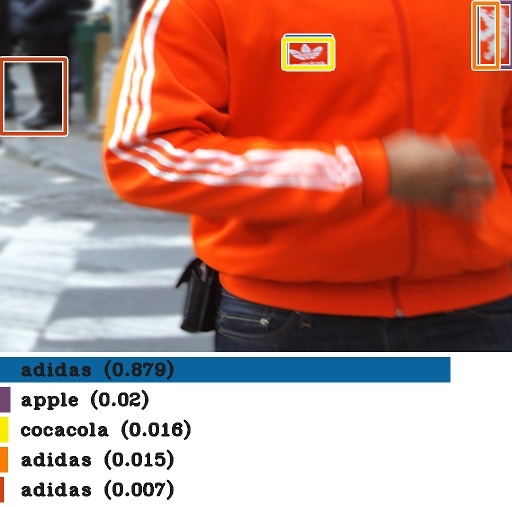}
\end{center}
   \caption{Top-5 confidence scores for an example image belonging to the Adidas class.}
\label{fig:esso}
\end{figure}

In Figure \ref{fig:top-5} we can observe that regions with a graphic logo are often assigned with a higher value of confidence than background regions. Most graphic logos tend to be simplistic and therefore contain less information than more complex ones. For example, the HP logo will blend more easily into background noise than the Starbucks logo since it has less characteristic features. To reduce this problem and subsequently the number of false positives, more training data is required. Given the fact that graphic logos are mostly presented as a flat object or imposed over other objects, this is a scenario where it would be possible to perform data-augmentation or procedurally generating new images for this purpose.
The softmax classifier will give a confidence value for each class, throughout all proposed regions. After reducing the set of classified regions using non-maximum suppression we will use the region with top confidence value to classify each image. As an example, Figure \ref{fig:esso} shows the top-5 confidence scores produced by this system for a test image.

Table \ref{table:map_comparison} compares our results with other referenced methods, more specifically in~\cite{romberg2013bundle}. It does also illustrate our results obtained with the approach from two sides: learning rate and selective search. In the former (i) learning rates are changed in the training and in the latter (ii) the idea is to generate window `proposals' in the pre-processing stage.  We can verify that having a higher base learning rate of $0.005$ is not beneficial for learning, since both models are overfitting at $10000$ iterations, whereas starting with a learning rate of $0.001$ achieves higher testing performance from the beginning.
From the results, it is counter-intuitive that the Quality mode achieves worst performance as compared with the Fast mode. This could be explained by the difference of regions analyzed between the two modes. Although the Quality mode generates a larger and more diverse set of regions, they tend to be misclassified more often than in the Fast mode. 
In summary, by using the quality version variant of Selective Search, it does not lead necessarily to better performance. This might be explained due to a higher number of windows (that do not contain a graphic logo) that have to be evaluated.
\begin{figure*}[htb]
  \centering
  \includegraphics[width=0.7\textwidth]{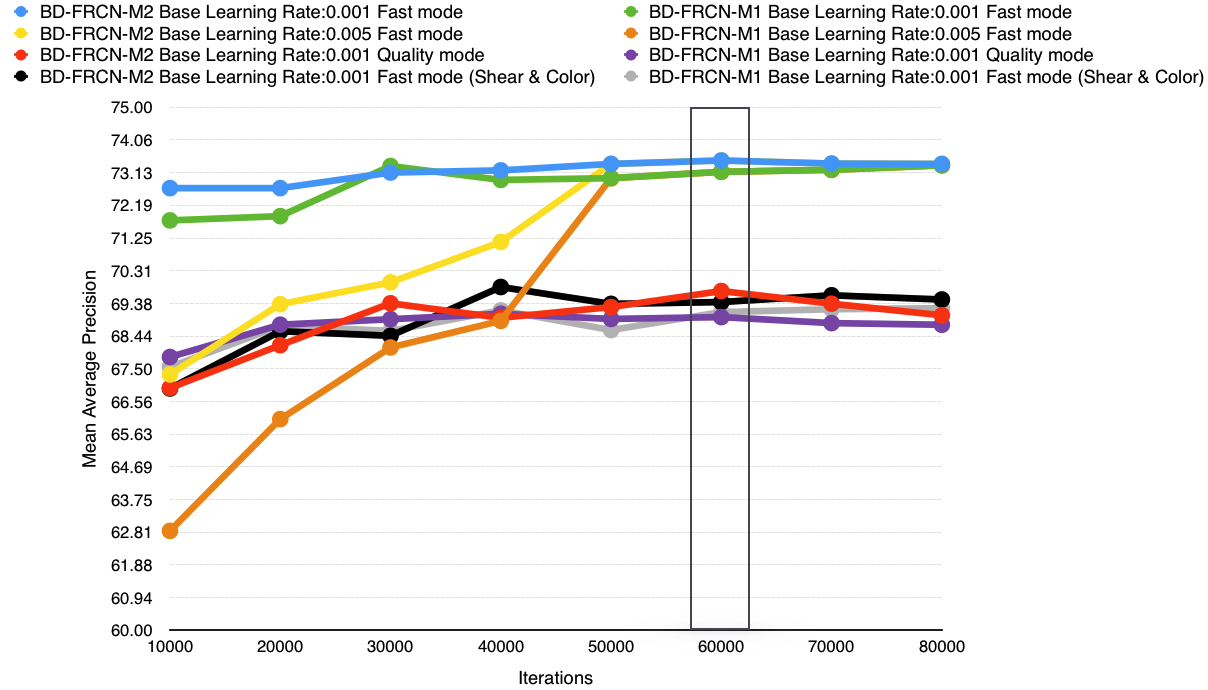}
  \caption{\ac{mAP} values across iterations with each network and several configurations.}
    \label{fig:map}
\end{figure*}
In Figure \ref{fig:map} we compare the test set \ac{mAP} values during the fine-tuning process for the two networks and several configurations. This means that the values across iterations with each network and configuration are indicated. The choice to measure test performance across iterations is mostly due to the fact that we lost validation data by joining the training and validation set for training. It is possible to achieve high \ac{mAP} if both precision and recall are high across detections.  As it can be seen, one of the configurations is performed with data augmentation.
\begin{figure}[h]
  \centering
 	\subfigure[Original]{\includegraphics[width=0.175\textwidth]{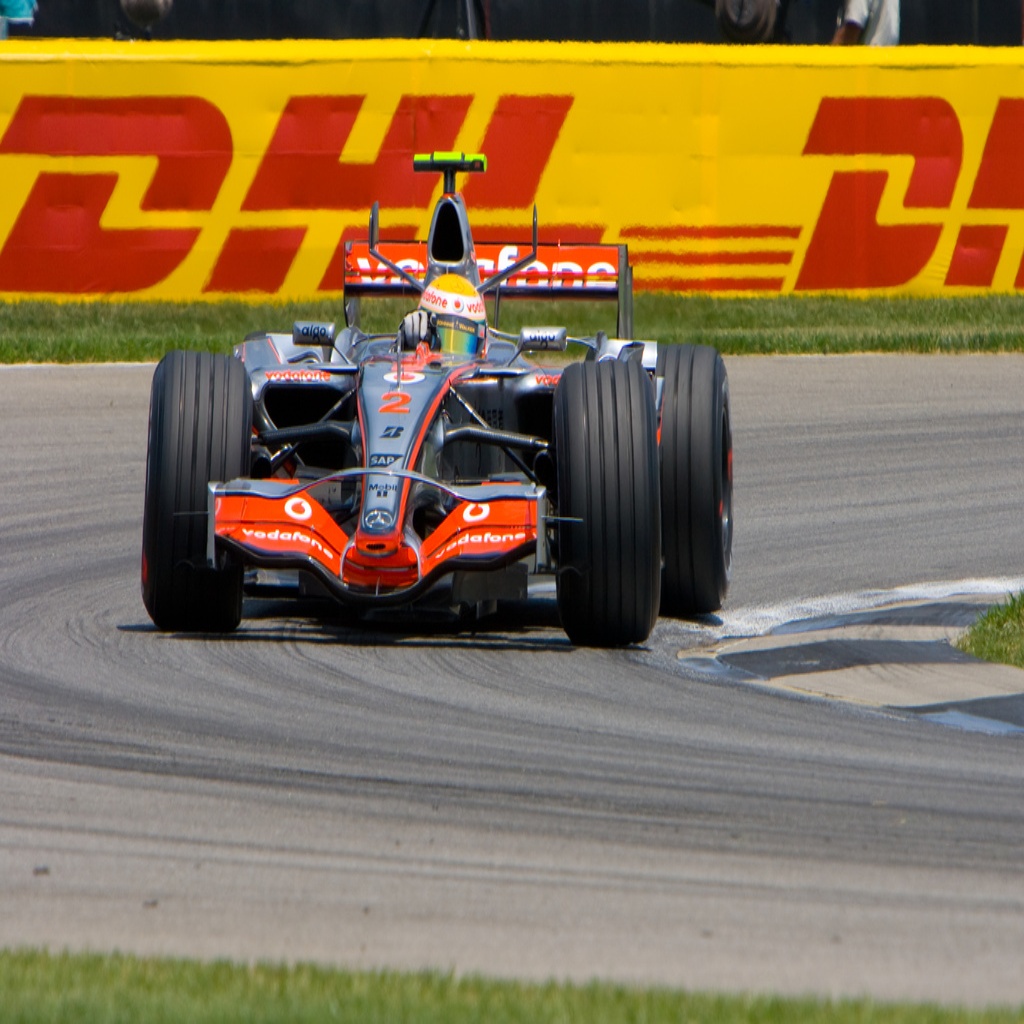} }
    \subfigure[Variation]{\includegraphics[width=0.175\textwidth]{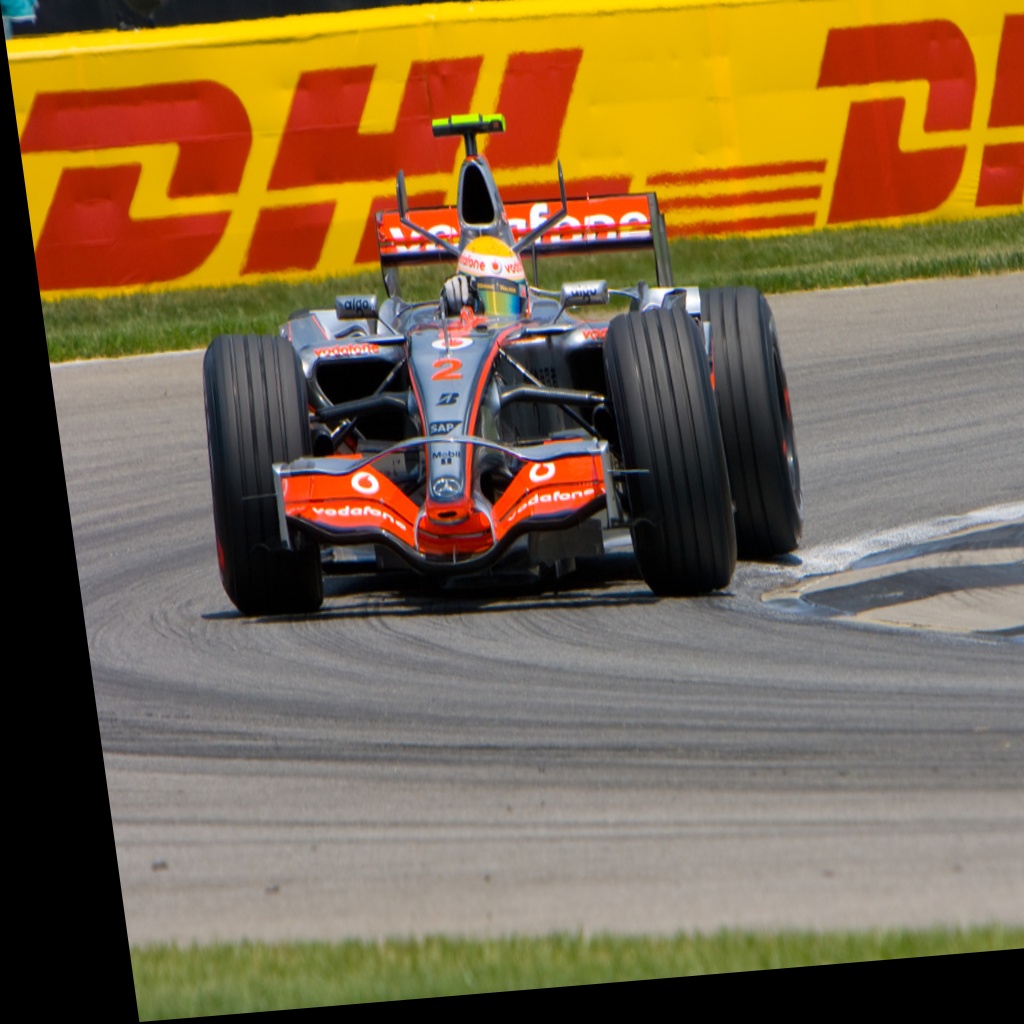} }
  \caption{Example of image variations produced.}
      \label{fig:variations}
\end{figure}
Additionally to the data augmentation performed by randomly flipping each region, we performed a small experiment by randomly distorting each image with a shear and slightly shifting color values on the image. We can observe a higher average precision for some classes, such as the Google, Guinness and BMW, but this fact is not reflected across all classes. In practice we doubled the number of images on our training set, by saving each variation. Further study on this aspect is required. Figure \ref{fig:variations} shows and example of such  variation.

Since our model classifies each image based on regions, and we do not have explicit confidence scores for the no-logo class, we use the region on the image with top confidence value and a threshold value to infer if only background regions are present. If the top confidence value is below a certain threshold value, we classify that image as a no-logo image, instead of the predicted logo class.
\begin{figure}[!]
\begin{center}
\includegraphics[width=0.375\textwidth]{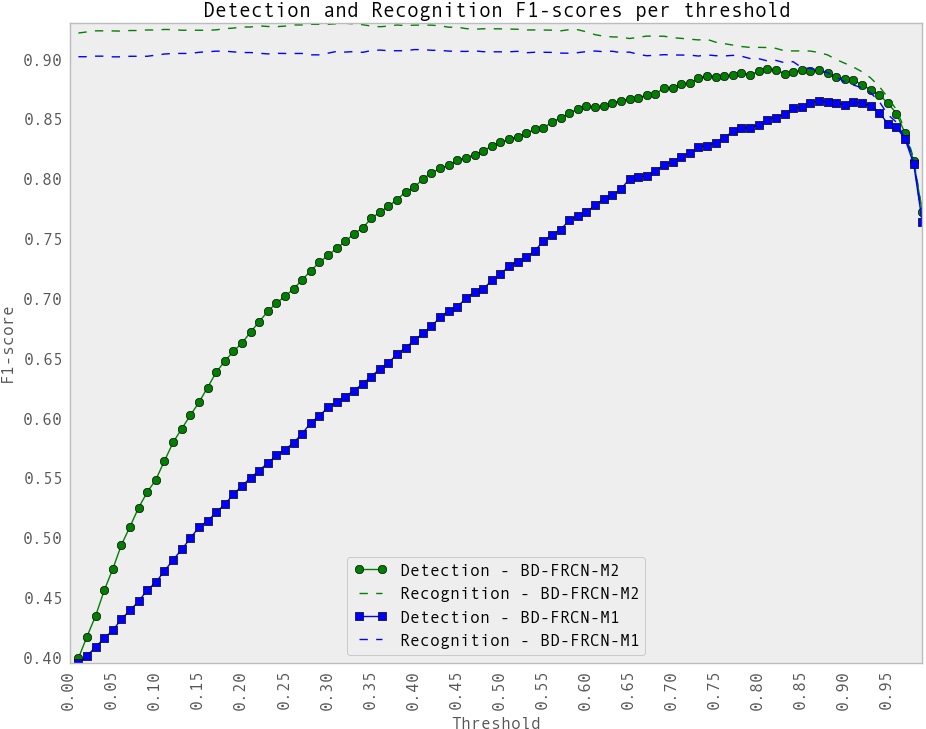}
\end{center}
   \caption{F1-scores for both BD-FRCN-M$_1$ and BD-FRCN-M$_2$.}
\label{fig:thresholds}
\end{figure}
 \begin{figure}[!]
  \centering
\subfigure[Class: no-logo]{{\includegraphics[width=0.20\textwidth]{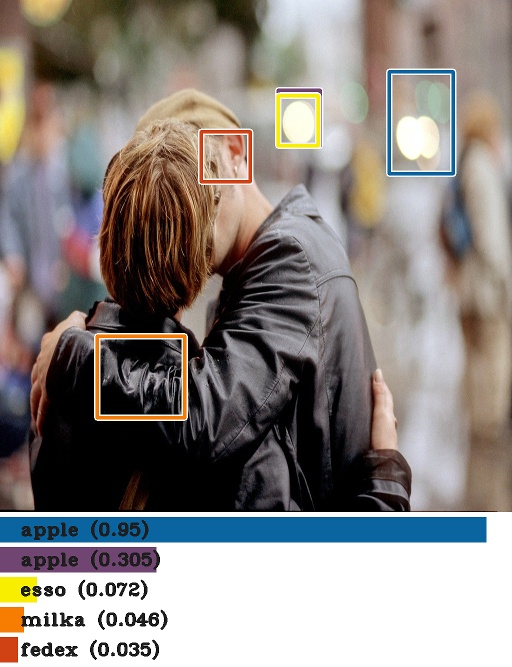} }}  ~ 
\subfigure[Class: no-logo]{{\includegraphics[width=0.20\textwidth]{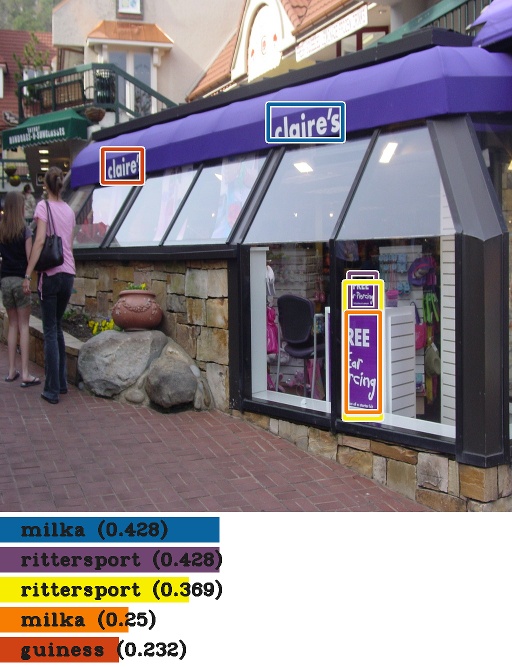} }}  ~            
  \caption{Examples of incorrect detections in the no-logo class.}
      \label{fig:no_logo_examples}
\end{figure}
Introducing the $3000$ no-logo images from the test set, we computed recognition F1-scores for both networks across learning rates iterations and configurations, for each threshold value on a range 0-1.
So far we have assumed that all images contain a graphic logo. We introduce a threshold value to deal with images without graphic logos (non-logo). Images with top confidence values below the threshold will be considered as having no logo present. Figure \ref{fig:thresholds} shows consolidated F1-scores for both detection and recognition metrics and the two tested CNN models. Using a threshold value of $0.3$ we achieve the top recognition F1-score, although by setting an adequate threshold value still allows the model to maintain high recognition scores without losing the ability to perform detection between images with logos and no graphic logos (\textit{e.g.} threshold $0.8$). 
We can observe in Figure \ref{fig:thresholds}, both metrics interacting. As we set higher threshold values the model starts to correctly identify more no-logo images and the detection score rises, at the same time, recognition scores start lowering since we start to incorrectly classify logo images as having no-logo. At high threshold values both scores start to go on a downward trend, since few top confidence values reside on that range.
 
Our model is able to distinguish most of the background regions from logos. However under specific conditions ends up giving high confidence values for regions that do not contain a logo from the list of 32 classes.
Figure \ref{fig:no_logo_examples} show some of those images, for example, the Apple and Milka logo are often confused for white backgrounds and purple backgrounds, respectively. Graphic logos, that are present in the images and do not belong to the $32$ classes are also sometimes wrongly detected.
 
 Table \ref{table:values} also shows the state-of-the-art results achieved by other authors on this specific dataset that we know of, and as we can see, our results with this model are better, specifically if we look at the F1 score. 
The top recognition F1-score of $0.931$ was found using the BD-FRCN-M$_2$ model, with a base learning rate of $0.001$ at $40000$ iterations and a threshold value of $0.32$.
The BD-FRCN-M$_1$ model achieved a top recognition F1-score of $0.909$ with a base learning rate of $0.001$ at $30000$ iterations and with a threshold of $0.4$.
\begin{table}[thb]
\centering
\caption{Recognition scores}
\begin{tabular}{l|l|l|l}
\hline
Method                         & Precision & Recall    & F1        \\
\hline \hline
Brugman AlexNet-logos-3000~\cite{brugman}     & 0.713     & 0.569     & 0.633     \\
Brugman NIN-logos-3000~\cite{brugman}  & 0.705     & 0.604     & 0.651     \\
Brugman Caffenet-logos-3000~\cite{brugman}    & 0.729     & 0.565     & 0.637     \\
Romberg et al.~\cite{RombergICMR2011} & 0.98      & 0.61      & 0.752     \\
Romberg and Lienhart~\cite{romberg2013bundle} & 0.99      & 0.832     & 0.904     \\
BD-FRCN-M$_1$ (thresh 0.4)   & \textbf{0.928} & \textbf{0.891} & 	\textbf{0.909} \\
BD-FRCN-M$_2$ (thresh 0.81)  & \textbf{0.987} & \textbf{0.846} & 	\textbf{0.911} \\
BD-FRCN-M$_2$ (thresh 0.32)  & \textbf{0.955} & \textbf{0.908} & 	\textbf{0.931} \\
\end{tabular}
      \label{table:values}
\end{table}

\begin{figure}[h]
  \centering
 	\subfigure[Class: Tsingtao]{{\includegraphics[width=0.175\textwidth]{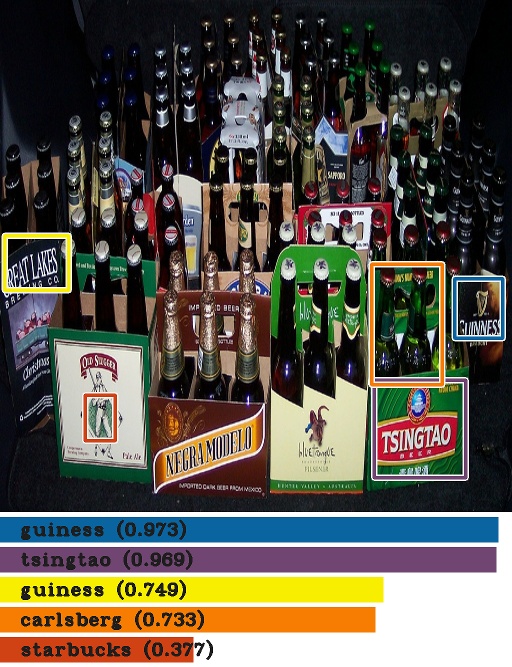} }}  ~
    \qquad
    \subfigure[Class: Heineken]{{\includegraphics[width=0.175\textwidth]{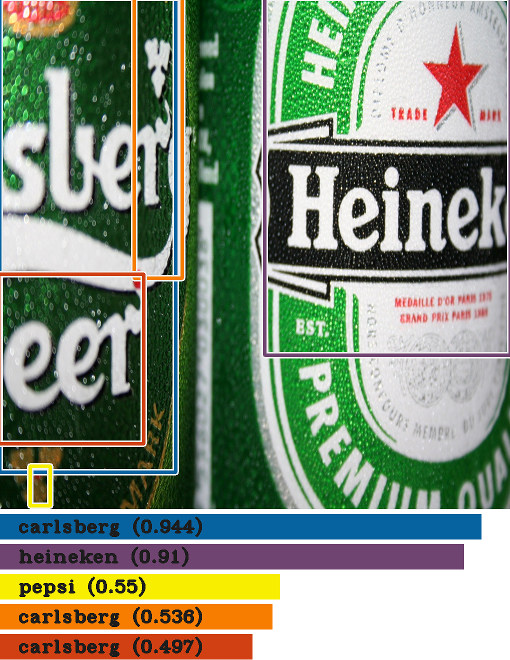} }}  ~ 
  \caption{Examples of two different graphic logos in the image}
      \label{fig:incongruences}
\end{figure}

As a side note, the model was able to find some inconsistencies in the dataset. It is stated that no two graphic logos are present in the same image. However, Figure \ref{fig:incongruences} shows two examples where the two top windows are correctly assigned to the class, but the top detection does not correspond to dataset ground truth.
\section{Conclusion}
\label{sec:conclusion}
A key contribution of this work has been the introduction of graphic logo detection using regions `proposals' and CNNs, by taking advantage of transfer learning.  
We experimented with a modern detection model and two CNNs pre-trained for a general object recognition task with abundant data and fine-tuned these networks for a task where the data is scarce - graphic logo detection -. 
Both  proposed models perform better than those of the state-of-the-art performance almost out of the box, with a wide margin for improvement that we intend to explore further.
In the future, we will  extend this model by exploring specific ways to generate regions where the presence of a graphic logo is probable, instead of using a class agnostic method like Selective Search. We are also planning on exploring data augmentation techniques to generate realistic transformations of graphic logos. Saving each image variation is not feasible due to disk space constraints, so implementing a module that records this data during training is further required. Another future issue will be the identification of brand logos from video data.
\bibliographystyle{IEEEtran}
\bibliography{egbib,Bibliography}
\end{document}